\documentclass[letterpaper, 10 pt, conference]{ieeeconf}  

\IEEEoverridecommandlockouts                              

\usepackage{graphicx}
\usepackage{verbatim}
\usepackage{array}
\usepackage{upgreek}
\usepackage{float}   
\usepackage{amsmath,amssymb}
\usepackage{cite}
\usepackage{booktabs}
\usepackage{algorithm}
\usepackage[noend]{algpseudocode}

\usepackage{todonotes}
\usepackage{xcolor}

\makeatletter
\def\algbackskip{\hskip-\ALG@thistlm}
\makeatother

\usepackage{tikz}
\usetikzlibrary{positioning, shapes, arrows.meta}
\usetikzlibrary{shapes,arrows,chains}
\usetikzlibrary{arrows,calc,automata}

\tikzset{
    block/.style = {draw, rectangle, 
        minimum height=.5cm, 
        minimum width=1cm},
    input/.style = {coordinate,node distance=.5cm},
    output/.style = {coordinate,node distance=.5cm},
    arrow/.style={draw, -latex,node distance=.5cm},
    pinstyle/.style = {pin edge={latex-, black,node distance=1cm}},
    sum/.style = {draw, circle, node distance=1cm},
    line/.style={-{Stealth}}
    }

\title{\LARGE \bf
Empirical Study of Ground Proximity Effects \\ for Small-scale Electroaerodynamic Thrusters
}

\author{
Grant Nations$^{1}$, C. Luke Nelson$^{2}$ and Daniel S. Drew$^{3}$ 
\thanks{$^{1}$Kahlert School of Computing, University of Utah, Salt Lake City, UT 84112, USA}
\thanks{$^{2}$Department of Mechanical Engineering, University of Utah, Salt Lake City, UT 84112, USA}
\thanks{$^{3}$Department of Electrical and Computer Engineering, University of Utah, Salt Lake City, UT 84112, USA}
\thanks{Corresponding author: Daniel S. Drew, \tt{daniel.drew@utah.edu}}
}

\begin{document}

\maketitle
\thispagestyle{empty}
\pagestyle{empty}

\begin{abstract}
Electroaerodynamic (EAD) propulsion, where thrust is produced by collisions between electrostatically-accelerated ions and neutral air, is a potentially transformative method for indoor flight owing to its silent and solid-state nature. 
Like rotors, EAD thrusters exhibit changes in performance based on proximity to surfaces. 
Unlike rotors, they have no fragile and quickly spinning parts that have to avoid those surfaces; taking advantage of the efficiency benefits from proximity effects may be a route towards longer-duration indoor operation of ion-propelled fliers.
This work presents the first empirical study of ground proximity effects for EAD propulsors, both individually and as quad-thruster arrays. 
It focuses on multi-stage ducted centimeter-scale actuators suitable for use on small robots envisioned for deployment in human-proximal and indoor environments. 
Three specific effects (ground, suckdown, and fountain lift), each occurring with a different magnitude at a different spacing from the ground plane, are investigated and shown to have strong dependencies on geometric parameters including thruster-to-thruster spacing, thruster protrusion from the fuselage, and inclusion of flanges or strakes. 
Peak thrust enhancement ranging from 300 to 600$\%$ is found for certain configurations operated in close proximity (0.2 mm) to the ground plane and as much as a 20$\%$ increase is measured even when operated centimeters away.
\end{abstract}

\section{Introduction}

The small size of micro air vehicles (MAVs) makes them well-suited to applications demanding operation in constrained environments, including cluttered and / or indoor inspection and mapping~\cite{fraundorfer_vision-based_2012, li_autonomous_2013, shen_autonomous_2011, quenzel_autonomous_2019}, cave and subsurface complex exploration~\cite{roucek_darpa_2020,iii_distributed_2017}, and warehousing~\cite{eudes_autonomous_2018,beul_autonomous_2017,kwon_robust_2020}. These contexts demand high levels of maneuverability and trajectory precision to avoid damage to the vehicle or the environment---or to people, if operated near them. At the same time, scaling challenges (e.g., with propeller aerodynamics, motor efficiency, and battery power density) make the achievable flight duration for MAVs prohibitively short, precluding their more widespread adoption~\cite{karydis2017energetics,mulgaonkar_power_2014}. Proximity to surfaces during flight affects the aerodynamic performance of all flying vehicles; there are a range of beneficial and deleterious effects depending on airfoil, actuator, and angle (e.g., if the surface is above, below, or to the side of the vehicle). Properly accounting for these effects is critical for designing robust and safe controllers, and they can have a potentially large impact on energy efficiency. 

\begin{figure}
    \centering
    \includegraphics[width=\columnwidth]{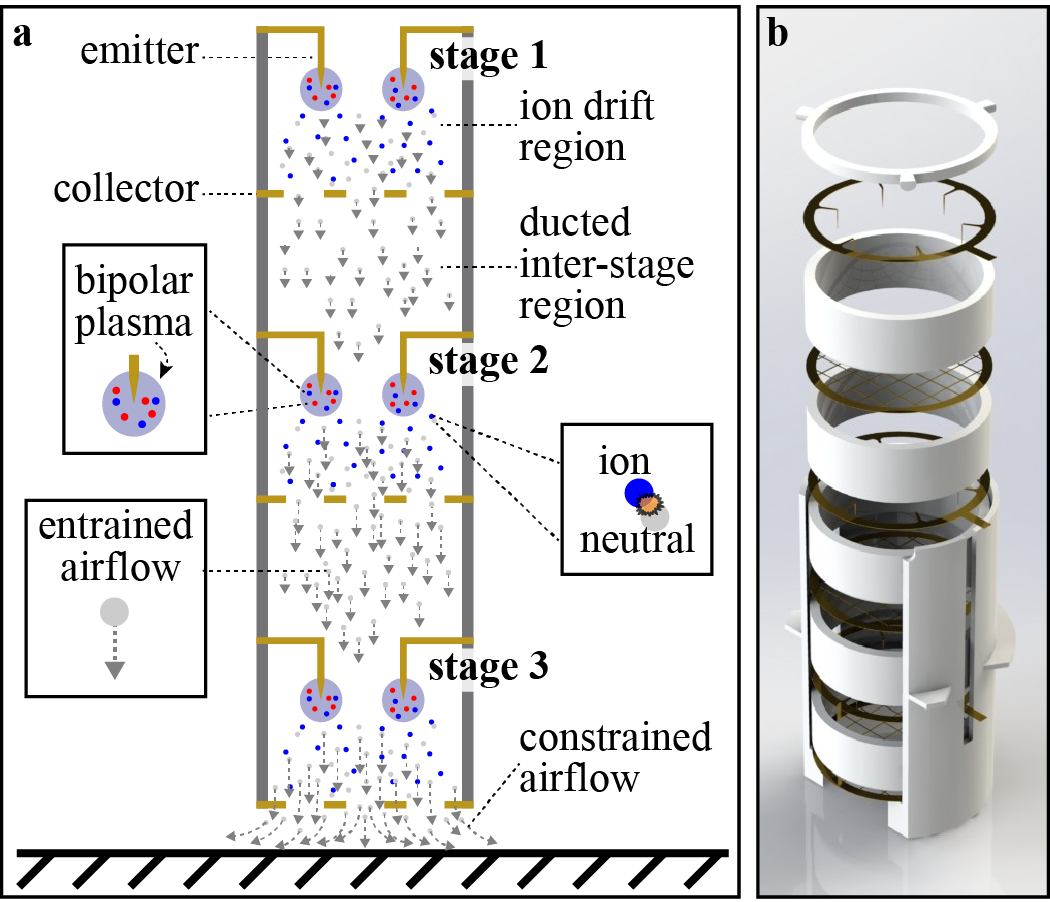}
    \caption{\textbf{(a)} Schematic view of a three-stage electroaerodynamic ducted actuator. A high potential applied between the ``emitter'' and ``collector'' electrodes ignites a local plasma from which ions are ejected. These ions drift in the electric field towards the collector, and frequent momentum-transferring collisions with neutral air molecules along the way result in an entrained air jet. Successive stages increase the velocity of the air as it passes through the duct. \textbf{(b)} Render of the three-stage ducted devices used in this work, where UV-laser micromachined brass electrodes are integrated with SLA-printed ducts in an adhesiveless process. Design, fabrication, and assembly of these devices is drawn from prior work~\cite{nelson_high_2023}.}
    \label{fig:schematic}
    \vspace{-2mm}
\end{figure}

Electroaerodynamic (EAD) propulsion, where thrust is produced through the momentum-transferring collisions between charged particles accelerated in an electric field and neutral air molecules (Fig.~\ref{fig:schematic}), is an attractive mechanism for MAVs due to its silent and solid-state nature. Several small (i.e., centimeter-scale) robots have already been demonstrated which take advantage of EAD (also known as electrohydrodynamic, or EHD) actuators~\cite{drew_toward_2018,zhang_low_2021,prasad_laser-microfabricated_2020,zhang_centimeter-scale_2022}. These robots are particularly well-suited for use in human-proximal environments in order to avoid the negative health and productivity outcomes caused by exposure to consistent loud noise, like that produced by rotorcraft~\cite{wojciechowska_designing_2019,goh_workplace_2015,costa_work_2003,schaffer_drone_2021,christian_initial_nodate}. As they have no mechanical moving parts, EAD actuators are also safer to operate close to surfaces as compared to fragile rotors or flapping wings.

\begin{figure}[t]
    \centering
    \includegraphics[width=\columnwidth]{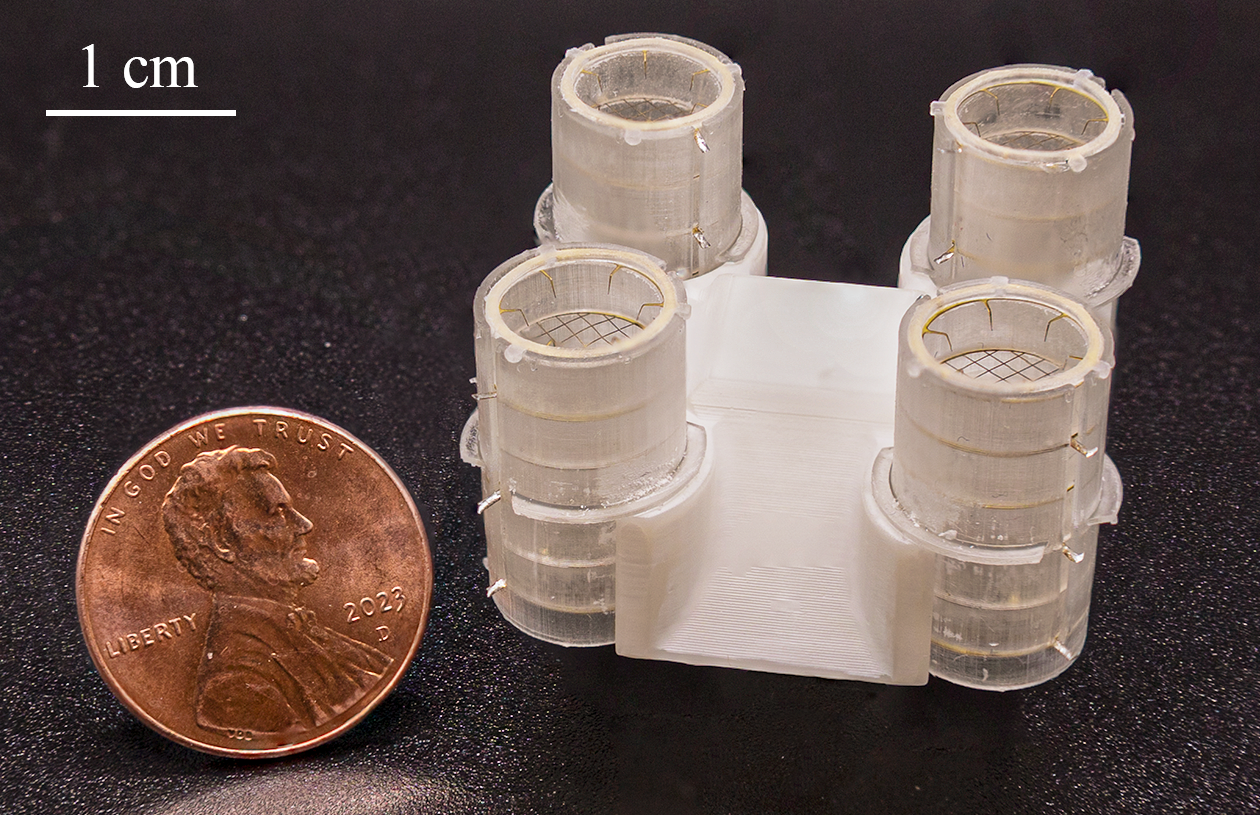}
    \caption{Four ducted thrusters arranged in a ``quadthruster'' configuration, with 45\textdegree~strakes extending down to form a skirt between them in order to minimize suckdown when in proximity to the ground. This design is shown to improve thrust by over 60$\%$ in close proximity to the ground, avoid any region of decreased thrust from suckdown, and still produce up to a 20$\%$ thrust benefit from fountain lift when further from the ground plane.}
    \label{fig:teaser}
    \vspace{-2mm}
\end{figure}

Despite a clear need, there has been no study of proximity effects as they apply to EAD propulsors to date. Here, we consider multi-staged ducted EAD thrusters suitable for small robots. We leverage our past work~\cite{drew_toward_2018,drew_high_2021,nelson_high_2023} that presents more in-depth analysis of thruster configuration and performance in order to focus more narrowly on ground proximity effects (i.e., the aerodynamic phenomena present when actuator exhaust is pointed at a nearby surface) for single- and quad-thruster arrangements with various configurations---Fig.~\ref{fig:teaser} is one example.

The primary contribution of this work is a first-ever empirical investigation of ground proximity effects for EAD thrusters. It includes experiments looking at the impact of factors including distance from the ground, Reynolds number, inter-thruster spacing, thruster protrusion from the fuselage, and inclusion of strakes on measured thrust and thrust efficiency. Our results serve as an initial confirmation of which ground proximity effects are of primary importance for EAD-propelled robots and point towards fruitful future directions for experimentation and model development. We show that ground effects can be a net efficiency gain or loss depending on arrangement and surface distance, and show that it can increase static thrust by as much as $600\%$ in specific configurations, or decrease it by as much as $20\%$ in others. The magnitude of this effect means that it should not be ignored in future studies of EAD-propelled MAVs, and opens the door to entirely new designs for high-efficiency vehicles (e.g., ion-propelled micro-hovercraft).

\section{Background and Related Work}
This section attempts to succinctly describe the mechanisms by which ground proximity affects aerodynamic performance in general terms before pointing towards relevant related work in the domain of MAVs. Information on electrohydrodynamic / electroaerodynamic ground proximity effects more specifically is not included because it has not yet been investigated in the academic literature. 

\subsection{Physical Basis of Ground Proximity Effects}
\begin{figure}
    \centering
    \includegraphics[width=\columnwidth]{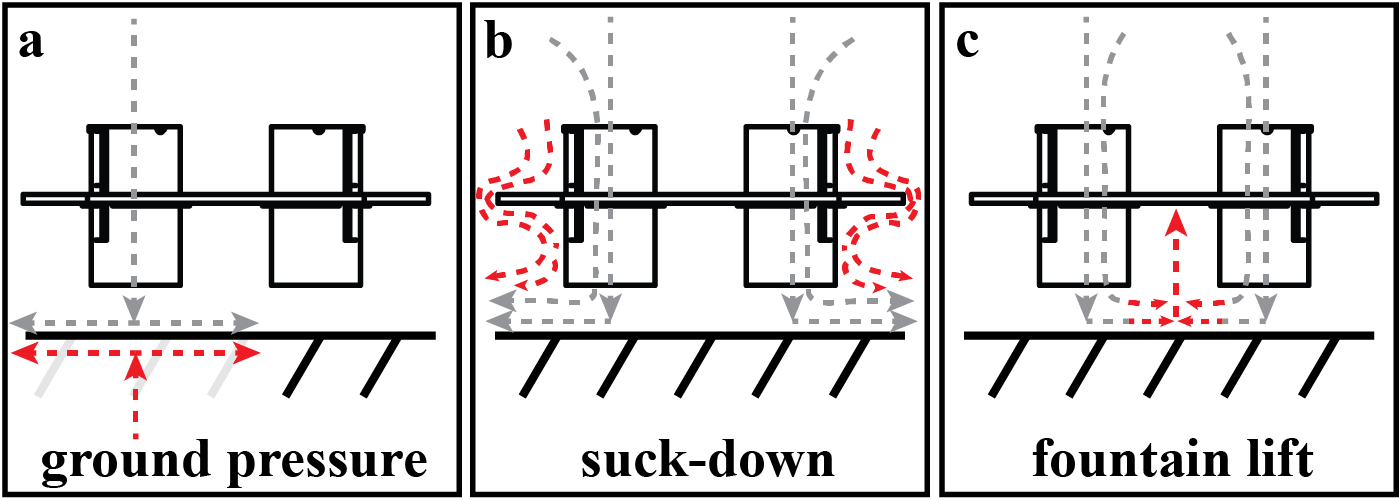}
    \caption{Relevant proximity effects include \textbf{(a)} the ground effect, where high pressure buildup from the exhaust causes a net thrust-positive reaction force, \textbf{(b)} suckdown, where the radial wall jet entrains and accelerates air flowing over the fuselage, and \textbf{(c)} fountain lift, where adjacent wall jets collide and redirect to push against the underside of the fuselage.}
    \label{fig:prox_mechanisms}
    \vspace{-2mm}
\end{figure}

As with most aerodynamic phenomena, ground proximity effects can be explained at several levels of complexity, and their specifics are closely tied to the physical embodiment (e.g., actuator type and placement, fuselage shape) of the vehicle. The three primary effects of interest are the pressure ground effect (what is often referred to as ``the'' ground effect), suckdown, and fountain lift (Fig.~\ref{fig:prox_mechanisms}). 

In general, the \textit{ground effect} refers to the apparent increase in thrust or lift experienced by an aircraft when in close proximity to the ground. It has been shown to affect fixed-wing, flapping-wing, rotory, and jet-powered vehicles across scales~\cite{conyers_empirical_2018,sanchez-cuevas_characterization_2017,mccormick_aerodynamics_1999,rozhdestvensky_wing--ground_2006, truong_aerodynamic_2013}. While varying explanations exist, those focused on airfoil tip vortex reduction and increased leading-edge streamline curvature, for example, fail to account for the ground effect shown in jet engines without airfoils. The method of images, where a virtual source is mirrored across the ground plane to represent ground reaction forces, is mathematically satisfying and has been shown to accurately predict lift~\cite{long_origin_2023}. Simply, increased proximity to the plane results in a more vertical center-duct streamline and higher velocity induced wall jet, which is reflected in the mirror image to produce a larger ground reaction force in the direction of lift. 

Proximity to a surface is not always beneficial; vertical and / or short takeoff-and-landing (V/STOL) aircraft also experience a negative effect known as \textit{suckdown}. This comes as a result of the vehicle's actuators entraining and accelerating air vertically downward and around the bottom edges of the aircraft (red streamlines shown in Fig.~\ref{fig:prox_mechanisms}b), creating an area of low pressure beneath the fuselage and a net ``suction'' pressure towards the plane~\cite{mccormick_aerodynamics_1999}. Viscous mixing between the entrained air and the jet exhaust also reduces jet thrust when the change in velocity is propagated through the streamline~\cite{bevilaqua_jet_2007}. 

Finally, in multi-rotor and multi-jet V/STOL aircraft, the wall jets produced by neighboring propulsors can collide and redirect beneath the aircraft to produce a vertical airflow which impacts the bottom of the fuselage and produces a net upward force known as \textit{fountain lift}. The magnitude of this effect is a complicated function of jet decay rate, inter-actuator geometry, and the ratio between wall distance and jet diameter (at least)~\cite{kuhn_hover_1987}. Fountain lift can also arise from reflections of individual thruster exhaust depending on collimation and angle-of-attack~\cite{yen_vertical_1981}.

\subsection{Ground Effects in Small Rotorcraft}
The most salient related work includes other empirical studies looking at proximity effects for small (i.e., centimeter-scale) rotors and rotorcraft. Besides the obvious difference of using EAD thrusters and not rotors, our work also notably investigates by far the smallest diameter propulsor of the studies we were able to find. In practice, this means that a low $z/r$ ratio for our devices is a proportionally lower actual distance from the ground plane.  

For example, Conyers et al. investigated the change in thrust in- and out- of ground effect for nine inch diameter rotors in isolation and in quadrotor arrangements~\cite{conyers_empirical_2018}, finding peak thrust enhancement of around 15$\%$ and 20$\%$, respectively, below a $z/r$ ratio of about unity. They also found evidence for fountain lift (with an approximately 5$\%$ benefit) for appropriately spaced quadthruster arrangements. Our work differs in that we show coarse control over fountain lift region location and magnitude via geometric parameters, and also investigate suckdown effects that are not apparent with their rotors. Dekker et al. investigated dual-rotor arrangements in close ground proximity, directly visualizing and confirming the presence of fountain flow~\cite{dekker_aerodynamic_2022}. A focus of their work was on predicting fountain flow re-ingestion by the propellers, which is not possible given the (relatively) long, ducted nature of our EAD thrusters. Jardin et al. used a single isolated microrotor propeller (radius 12.5 cm) to measure and model the ground effect, finding a peak thrust enhancement of approximately 75$\%$ in close proximity to the plane~\cite{jardin_aerodynamic_2017}. They also found a strong dependence on rotor pitch angle; although the analogous thruster angle of attack was not studied in this work, it is clear that it could be important. A review article by Matus-Vargas et al.~\cite{matus-vargas_ground_2021} contains many more examples of modeling and empirical efforts for rotorcraft in the ground effect, though primarily for much larger vehicles. 

Various authors have proposed quadrotor control architectures and / or physical embodiment which seek to take advantage of surface proximity to improve flight endurance~\cite{hsiao_ceiling_2019,ding_passive_2022,hsiao_energy_2023}. We view this as an exciting future direction for EAD-propelled robots.

\section{Methods}
\label{sec:methods}

\subsection{Thrusters}
The individual EAD thrusters used in this study are nearly identical to those recently presented by Nelson and Drew~\cite{nelson_high_2023}, with modifications made only to the mounting structure. A single thruster consists of three ducted stages (as shown in Fig.~\ref{fig:schematic}) with an inner diameter of 8 mm and a total height of 17.6 mm. Succinctly, duct structures are fabricated via stereolithographic (SLA) printing, active electrodes are UV-laser micromachined, and the components are integrated using mechanical affordances and an adhesiveless locking system; further design, fabrication, and assembly details are omitted here for brevity but can be found in~\cite{nelson_high_2023}. These thrusters have been found to have among the highest thrust densities recorded for any propulsor at this scale, with exhaust velocities approaching 5 m/s.

Inspired by prior work on centimeter-scale EAD-propelled robots~\cite{drew_toward_2018}, devices are assembled into ``quadthruster'' configurations using SLA-printed fuselage plates (Formlabs Form 3+, Rigid 10k resin) with printed locking clips (the latter are omitted from Fig.~\ref{fig:methods_combo} for clarity). Controlled geometric parameters, which will be referenced throughout subsequent sections, are labeled in Fig.~\ref{fig:methods_combo}a (left) and include the thruster radius, $r$, the inter-thruster spacing, $s$, the thruster protrusion from the fuselage, $h$, and the distance from thruster exhaust to the ground plate, $z$.

\subsection{Experimental Setup}
\begin{figure}
    \centering
    \includegraphics[width=\columnwidth]{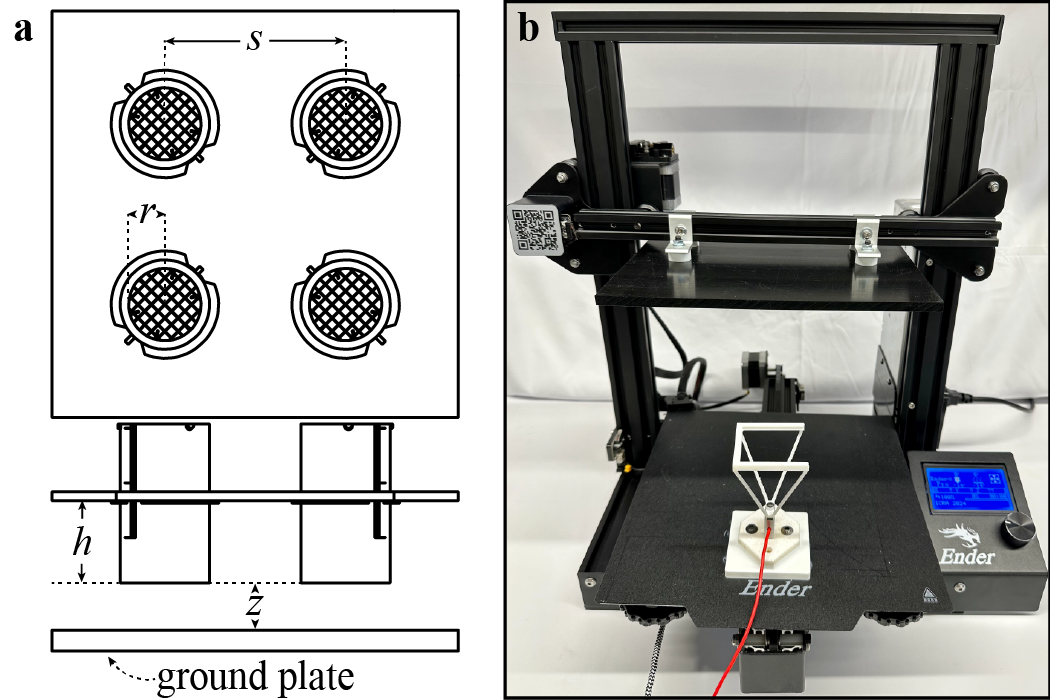}
    \caption{\textbf{(a)} Schematic illustrating geometric parameters of interest, including the inter-thruster spacing $s$, thruster radius $r$, thruster protrusion from the fuselage $h$, and the distance from the thruster exhaust to the ground plate $z$. \textbf{(b)} The automated test setup constructed from a modified 3D printer synchronizes current-voltage and force-voltage data acquisition with precise movement of the ground plate. An entire experimental suite (i.e., every desired $z/r$ value) can be initiated with a single command.}
    \label{fig:methods_combo}
    \vspace{-2mm}
\end{figure}

The automated experimental testbed is constructed from a modified Ender 3 Pro 3D printer (Fig.~\ref{fig:methods_combo}, right). The extruder and X-axis stepper are replaced with a 230 mm by 230 mm by 6.35 mm Delrin acetal plastic plate, secured to the extruded aluminum frame with countersunk M5 bolts. This plate acts as the ground, and is lowered or raised to the desired $z$ value using G-code commands sent from a Python script. The device under test is secured 53 mm above a FUTEK LSB200 S-Beam load cell in a rigid SLA-printed mount designed for unobstructed airflow, with thruster exhausts facing upwards towards the ground plate.

Voltage is applied to devices using a Spellman High Voltage SL8P programmable high-voltage power supply with a separate Python-controlled variable power supply connected to its remote inputs. Load cell readings are recorded in a separate thread at an interval of 100 mS. Voltages from the high voltage supply output monitor and from across a shunt resistor are recorded with a Rigol DS1054Z oscilloscope throughout each experimental trial (a voltage sweep up to $\approx$3kV) and used to calculate true applied voltage and output current. 
 
\section{Results and Discussion}
\label{sec:results}
All experiments are performed with two devices per test condition and three trials per device. Plotted data is the overall mean (mean of each device's trial mean) with standard error of the mean (SEM) error bars.

\subsection{Single thruster experiments}

\begin{figure}
    \centering
    \includegraphics[width=\columnwidth]{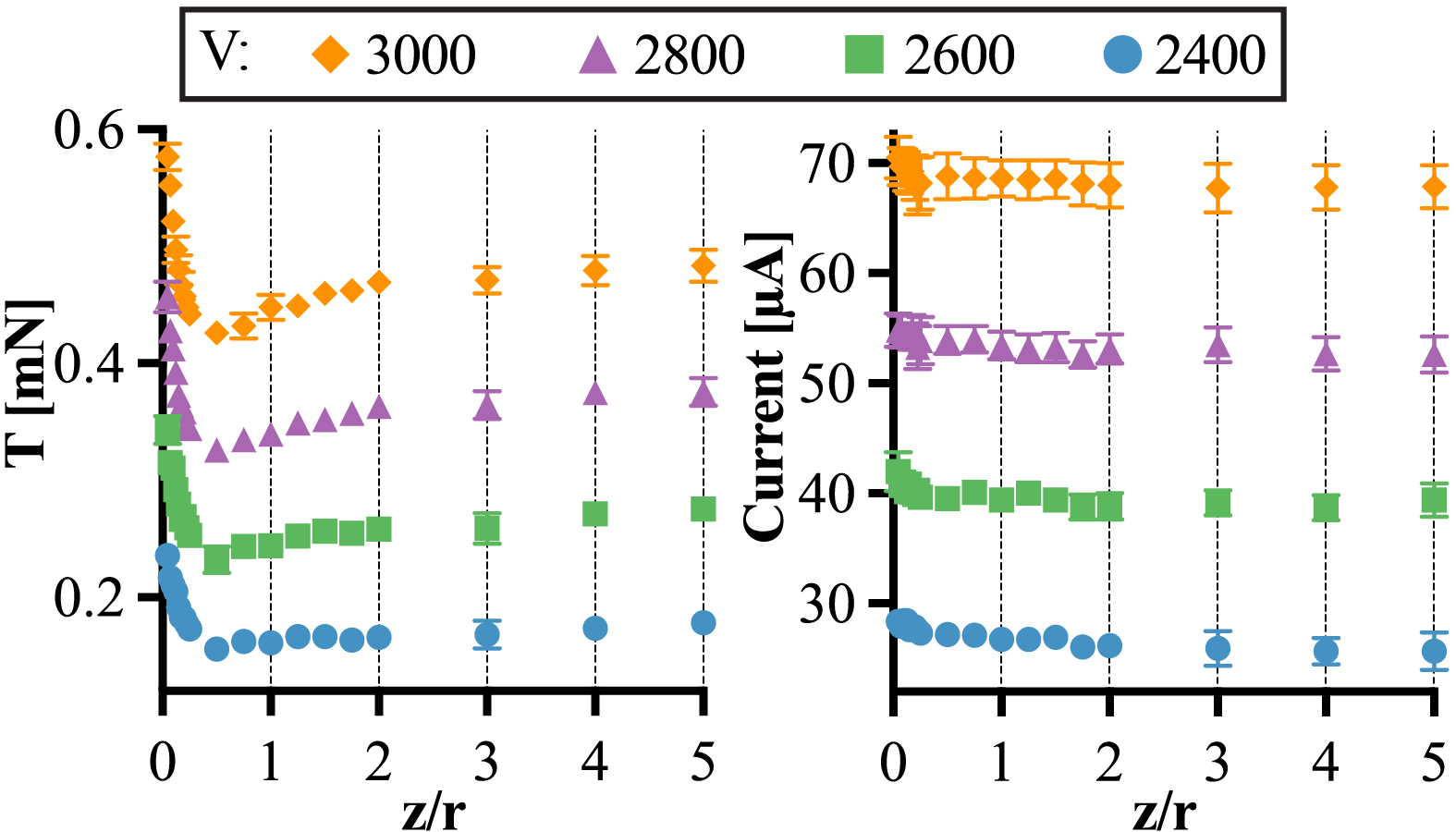}
    \caption{Thrust versus $z/r$ ratio (left) and current versus $z/r$ ratio (right) at different applied voltages for a single thruster.}
    \label{fig:thrust_dist_abs}
    \vspace{-2mm}
\end{figure}

\begin{figure}
    \centering
    \includegraphics[width=\columnwidth]{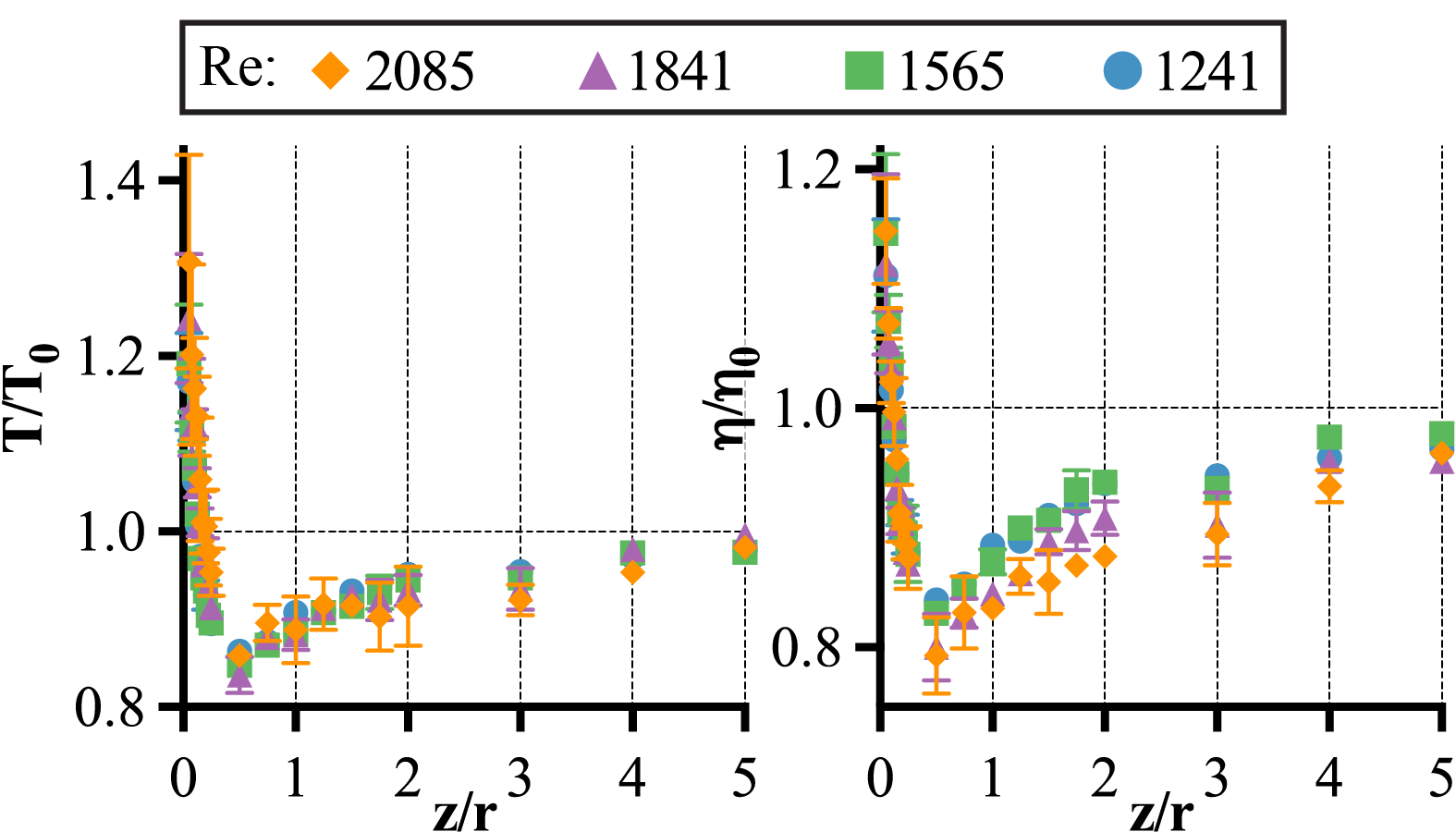}
    \caption{Ratio of measured thrust to out-of-ground-effect thrust ($z/r$ $>>$ 10) versus $z/r$ (left) and ratio of measured thrust efficiency to out-of-ground-effect thrust efficiency versus $z/r$ (right) for a single thruster at different Reynolds numbers, calculated based on force data and simple momentum theory.}
    \label{fig:thrust_re}
    \vspace{-2mm}
\end{figure}

\begin{figure}
    \centering
    \includegraphics[width=\columnwidth]{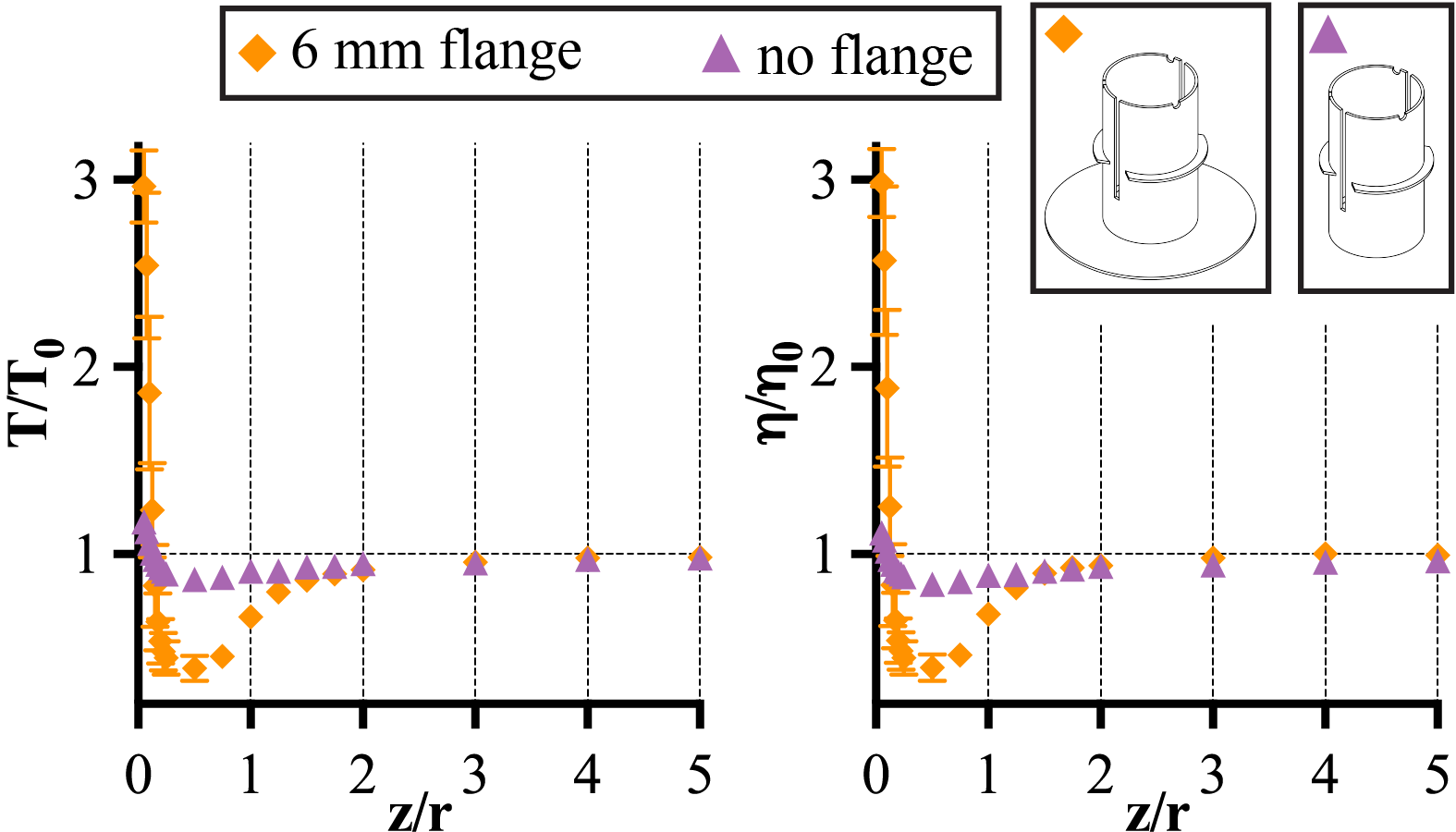}
    \caption{Ratio of measured thrust to out-of-ground-effect thrust versus $z/r$ (left) and ratio of measured thrust efficiency to out-of-ground-effect efficiency versus $z/r$ (right) for a single thruster with and without inclusion of an exhaust flange. All data collected at 3kV applied potential.}
    \label{fig:thrust_re_flange}
    \vspace{-2mm}
\end{figure}

Experiments for the ground effect of a single thruster consider the dependent variables of thrust ($T$), ion current, ratio of thrust to out-of-ground-effect (OGE) thrust ($T/T_0$), and ratio of thrust efficiency (in N/W) to OGE efficiency ($\eta/\eta_0)$, all with respect to $z/r$. 

As expected, the measured thrust increases with applied voltage as the ion-accelerating drift field increases in magnitude (Fig. \ref{fig:thrust_dist_abs}, left). Interestingly, the measured ion current at low $z/r$ ratios also increases (Fig. \ref{fig:thrust_dist_abs}, right) at the same applied voltage. We hypothesize that this is due to the increased magnitude of the convective term driving the ion current relative to the drift term; in ground effect, the increased velocity of the impinged wall jet should propagate through the device and increase inlet air velocity. For reference, the ion drift velocity (based on the formula $v_{drift} = \mu E$, where $\mu$ is the ion mobility and $E$ is the field magnitude, is expected to be roughly 100 m/s at reasonable operating voltages.  

Measured OGE thrust at different applied voltages can be used to calculate an approximate jet Reynolds number, where simple momentum theory is used to derive the average outlet air velocity via $T = 0.5 \rho A v^2$. We show that, at least over the limited achievable range of these devices, ground proximity effects do not have a strong dependence on Reynolds number (Fig.~\ref{fig:thrust_re}). It is unlikely that any realistic centimeter-scale EAD thruster will be able to reach much higher outlet velocities and Reynolds numbers (i.e., $> 10^4$) past the laminar-to-turbulent transition where we would definitely expect changes. 

Proximity to the ground results in up to a 30$\%$ increase in thrust relative to OGE thrust (Fig. \ref{fig:thrust_re}, left) at an applied voltage of 3 kV. In contrast with studies of isolated rotors, which never drop below the OGE thrust~\cite{conyers_empirical_2018,jardin_aerodynamic_2017}, the thrust then falls and remains below OGE thrust from $z/r \approx 0.25$ until $z/r \approx 5$. This is similar to measurements made for (much larger) isolated jets from VTOL aircraft, which see an anomalous amount of suckdown under-predicted by existing analytical models~\cite{kuhn_estimation_1991,bellavia_suckdown_1991}. 
As a result of the increased ion current at low $z/r$ ratios, the efficiency benefits are actually smaller than the thrust benefits (Fig.~\ref{fig:thrust_re}, right) by 10-20$\%$. This effect disappears at higher $z/r$ ratios, which supports the hypothesis that it is due to a proximity-induced increase in the convective current term.

Also considered is the effect of a circular flange extended 6 mm from the base of the thruster on the relative thrust and relative efficiency (Fig.~\ref{fig:thrust_re_flange}). Our hypothesis, which is that the increased surface area of the flange will increase benefits at low $z/r$ ratios due to an increased ground reaction force, but increase suckdown at higher $z/r$ ratios due to increased fuselage area for suction pressure to act on, is supported by the data. In this case the relative importance of the hypothesized ion convection term is harder to see relative to the ground effect; the relative efficiency is identical to within measurement deviation. 

\subsection{Multi-thruster experiments}

\begin{figure}
    \centering
    \includegraphics[width=\columnwidth]{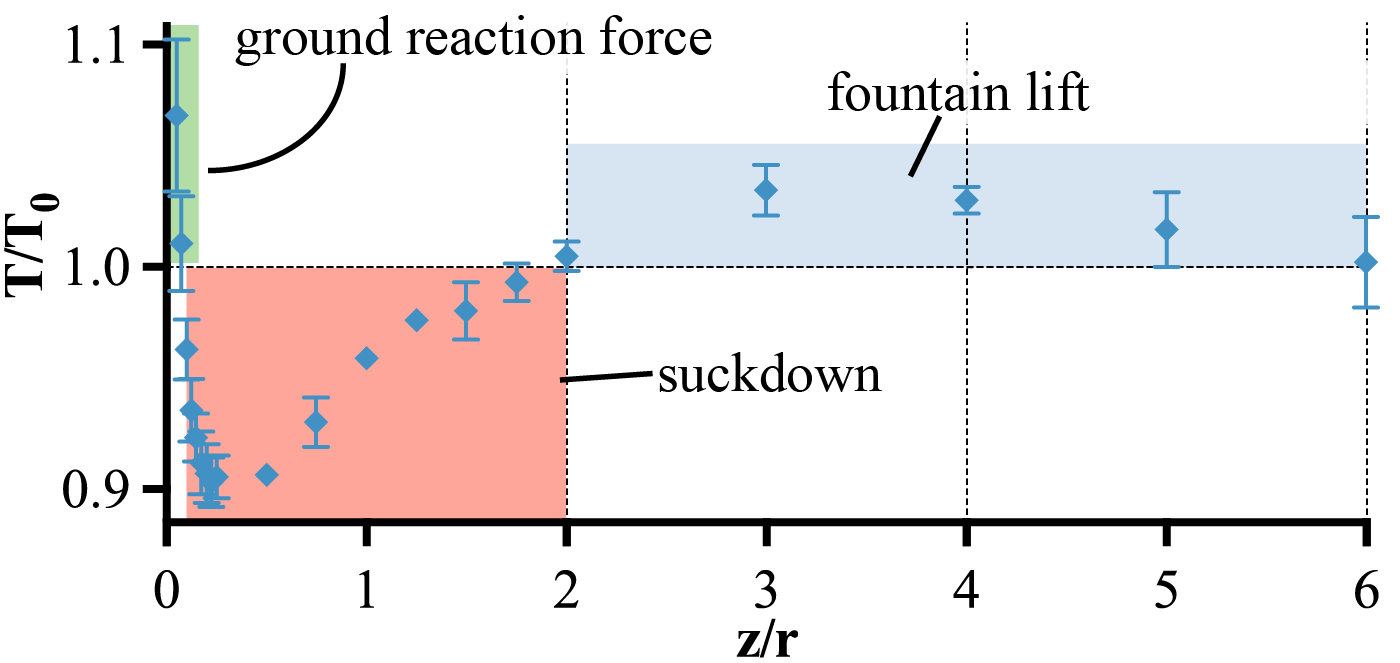}
    \caption{Quadthruster data ($s$ = 12 mm, $h$ = 9 mm, 3kV applied potential) with hypothesized regions of dominant proximity effect shaded and labeled. The precise location and size of these regions is expected to shift depending on the specific thrusters and multi-thruster geometric configuration.}
    \label{fig:quadthruster_regions_labeled}
    \vspace{-2mm}
\end{figure}

\begin{figure}
    \centering
    \includegraphics[width=\columnwidth]{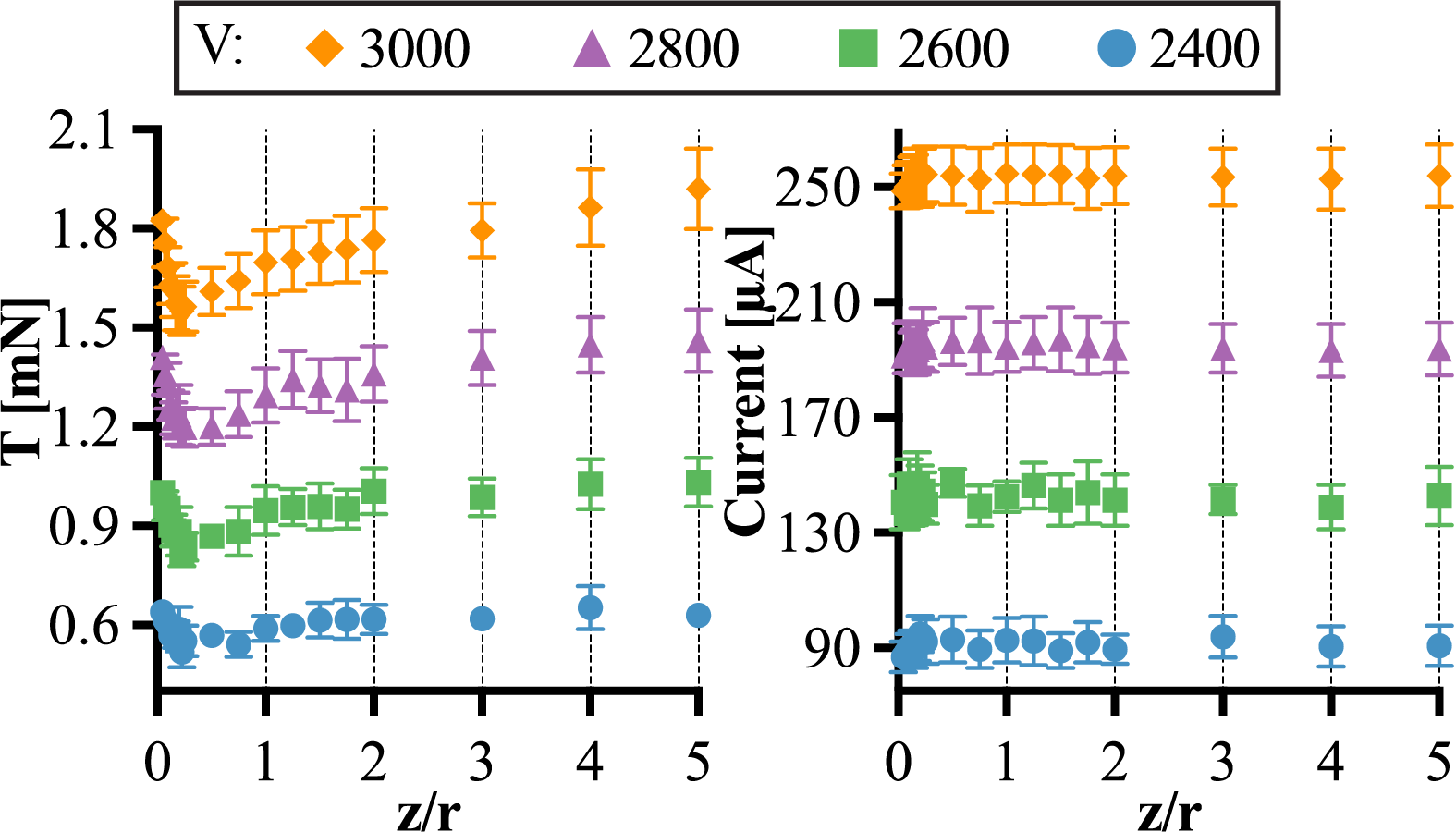}
    \caption{Thrust versus $z/r$ ratio (left) and current versus $z/r$ ratio (right) at different applied voltages for a quadthruster with $h$ = 9 mm and $s$ = 20 mm.}
    \label{fig:quad_thrust_dist_abs}
    \vspace{-2mm}
\end{figure}

\begin{figure}
    \centering
    \includegraphics[width=\columnwidth]{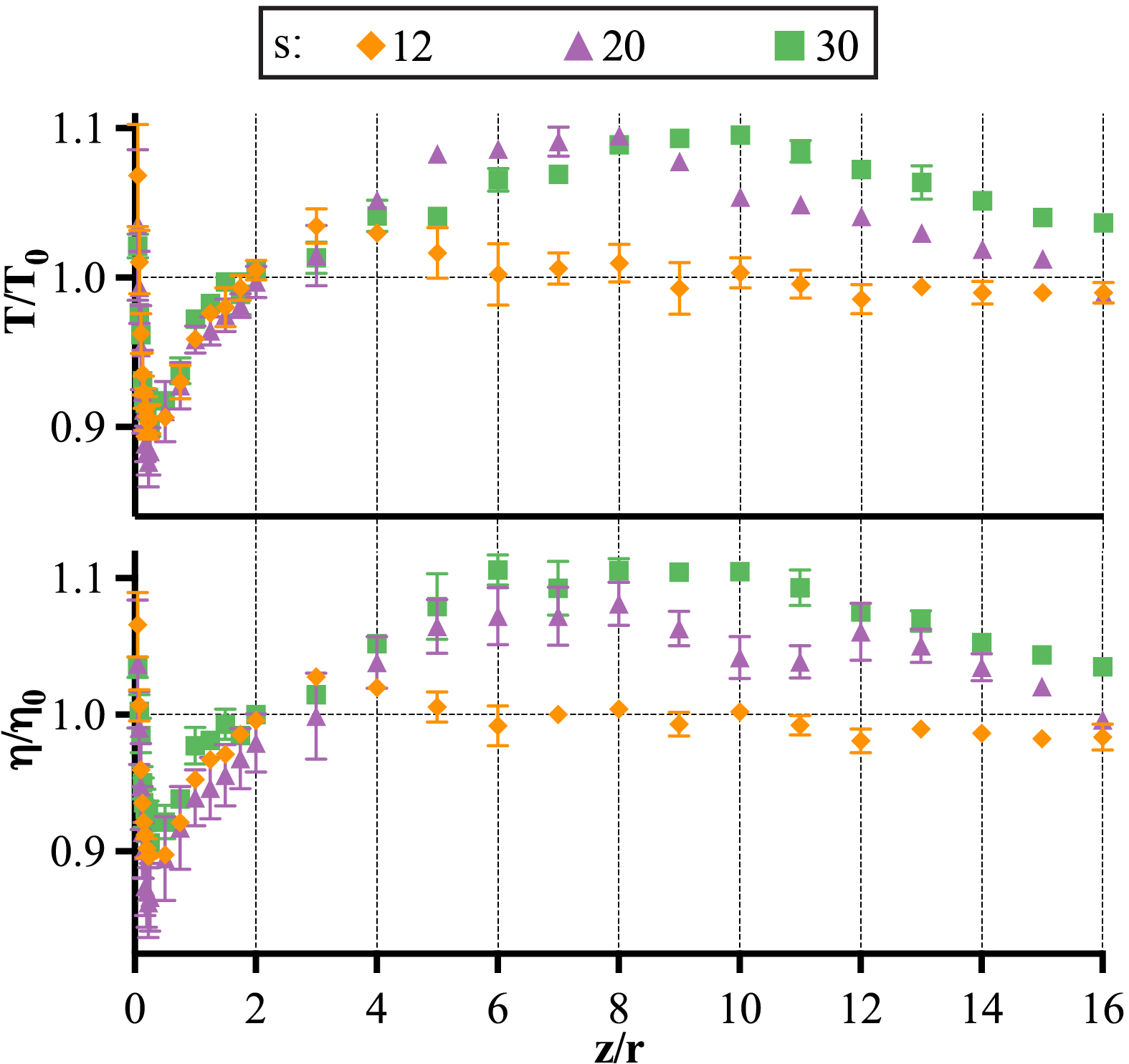}
    \caption{Ratio of measured thrust to out-of-ground-effect thrust versus $z/r$ for a quadthruster configuration with different inter-thruster spacing, $s$. All data collected at 3kV applied potential with $h$ = 9 mm.}
    \label{fig:quadthruster_s_relative}
    \vspace{-2mm}
\end{figure}

\begin{figure}
    \centering
    \includegraphics[width=\columnwidth]{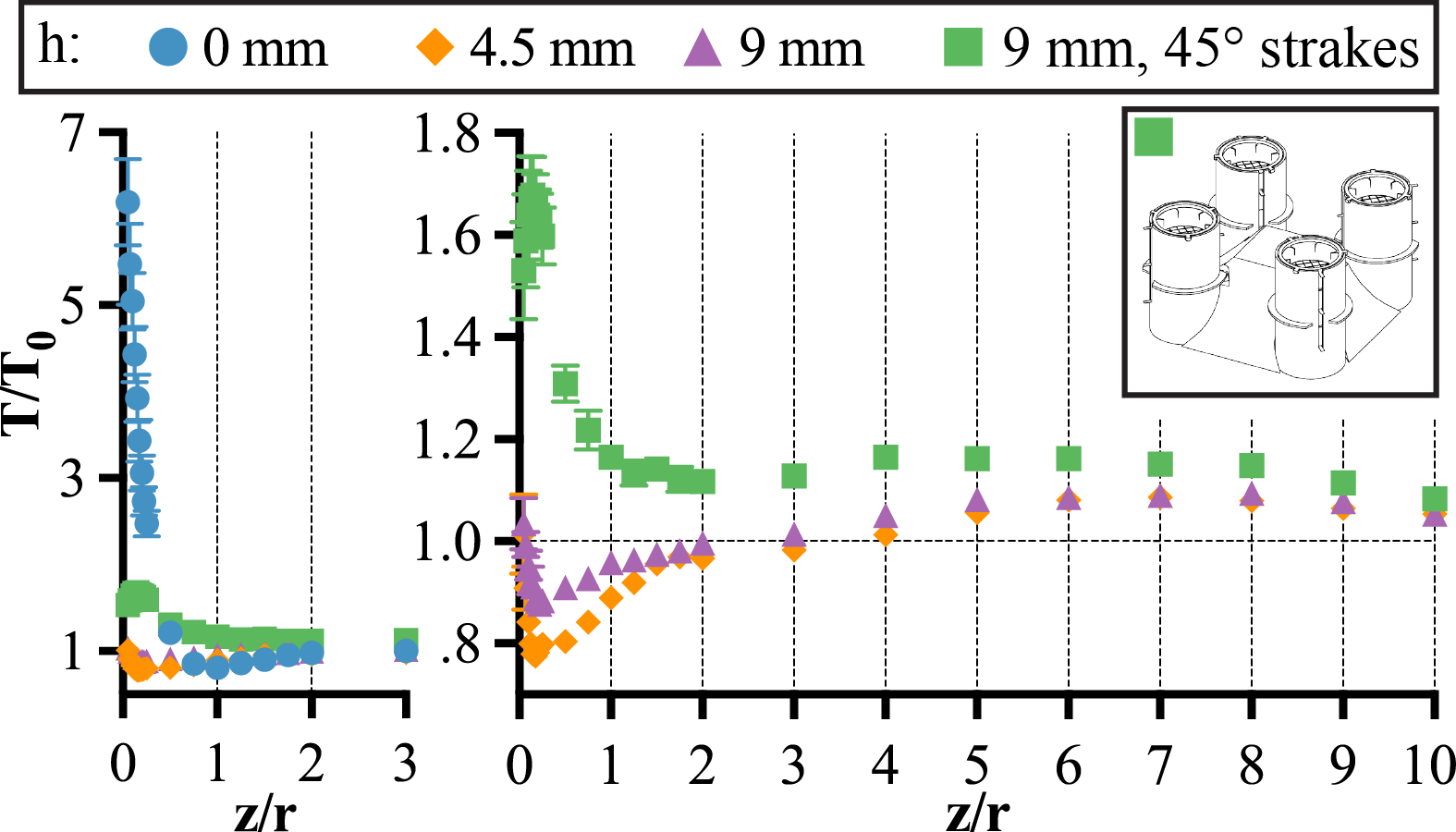}
    \caption{Ratio of measured thrust to out-of-ground effect thrust versus $z/r$ for a quadthruster configuration with different thruster protrusion heights, $h$, or inclusion of strakes (right). The same, but including $h$ = 0 data, which significantly skews the range of the measured thrust enhancement (left). All data collected at 3kV applied potential.}
    \label{fig:quadthruster_h_relative}
    \vspace{-2mm}
\end{figure}

We also tested quadthruster configurations with varying inter-thruster spacing, $s$, thruster protrusion, $h$, and inclusion of aerodynamic strakes. For a visual reference of these parameters see Fig.~\ref{fig:methods_combo}, left. Now that multiple jets are present, fountain lift manifests as the dominant effect at high $z/r$ ratios. All three (ground, suckdown, fountain) of the observed proximity effects are labeled with their dominant regions on top of a representative dataset in Fig.~\ref{fig:quadthruster_regions_labeled}.

Figure~\ref{fig:quad_thrust_dist_abs} shows the measured thrust and current for a quadthruster at different applied voltages. Unlike the data for the single thruster, we do not see an increase in current at low $z/r$ ratios, or if there is one it is within the measurement error. One hypothesis is that the impinged wall jet velocity is decreased from a combination of mixing with the entrained fuselage airflow and mixing with neighboring jet exhausts, and as a result the convective ion flow term no longer becomes significant. This is a complicated effect requiring direct flow visualization to assess.

The inter-thruster spacing was shown to have a strong effect on the location and magnitude of the thrust enhancement attributable to fountain lift (Fig.~\ref{fig:quadthruster_s_relative}). A possible explanation for why fountain lift is shown to continuing increasing until higher $z/r$ ratios for increasing values of $s$ is that those devices present a larger apparent surface area to the inner region where the exhaust streams intersect and are redirected upwards. The fountain jet is unlikely to be well collimated, as it has lost energy during redirection; much of the rising fountain lift airflow will just be pushed back to the ground plane by the jet exhausts at low spacing values. Importantly, the magnitude of the ground effect at very low $z/r$ ratios is shown to never exceed 10$\%$ here, compared to about 30$\%$ for the single-thruster case. We hypothesize that this due to the increased magnitude of the suckdown effect due to the larger fuselage area of a quadthruster. 

The magnitude of the suckdown effect relative to ground effect at low $z/r$ ratios showed a strong dependence on thruster protrusion from the fuselage (Fig.~\ref{fig:quadthruster_h_relative}). We hypothesize that suckdown is worse for the lower tested $h$ value because the streamline around the fuselage is closer to the impinged wall jet, so it is entrained and accelerated more strongly. When $h$ reaches 0 mm, however, there is no longer any space for the fuselage streamline to enter the underside of the device before being pushed out by wall jet, therefore there can be no force from a pressure differential across the fuselage (Fig.~\ref{fig:quadthruster_h_relative}, left). The magnitude of the ground effect also rises dramatically when $h=0$, although it is impossible to definitively say what percentage of that change is from removal of suckdown. This increase makes sense; the flat fuselage provides a larger surface area for the ground reaction pressure to act on, whereas at higher $h$ values the fuselage is at effectively a higher $z/r$ ratio than the thrusters. 

Aerodyamic slats known as strakes are commonly used on the underside of V/STOL aircraft to reduce the effect of suckdown (among other benefits)~\cite{mccormick_aerodynamics_1999}. We show that surrounding the fuselage with 45\textdegree$\,$ strakes which extend down to be flush with the thruster exhausts (Fig.~\ref{fig:teaser}) removes any visible suckdown effect, just like for the $h=0$ case (Fig.~\ref{fig:quadthruster_h_relative}, right); presumably, the streamlines follow the strakes and are unable to enter the underside of the device before being pushed out by the impinged wall jet. This data also supports our hypothesis for the magnitude of the ground effect in the $h=0$ case, because the rest of the fuselage is still elevated and a much more modest performance benefit is seen at low $z/r$ ratios. Interestingly, this is the only configuration which never shows a net performance decrease at any distance from the ground plane. 



\section{Limitations and Future Work}
While this study shows how various geometric parameters are useful for predicting and controlling ground proximity effects, this is a high-dimensional space and the physical phenomena are highly sensitive; a more rigorous experimental suite exploring the combinatorial geometric factors, for example using statistically-powered multi-factor analysis, is warranted in the future.

There are also several potentially interesting device parameters not explored here, including thruster radius, forward flight speed (or ambient flow velocity), and angle of attack. The maximum achievable and maximum change in Reynolds number were also limited by the operating range and number of stages in the devices. As related studies of rotorcraft and V/STOL aircraft note some dependence on all of these variables, we expect them to be important in future work.

Finally, our hypotheses about dominant proximity effect in each $z/r$ region are supported by the data, but without direct flow measurement are impossible to fully confirm. Flow visualization using particle image velocimetry (PIV) or another similarly high-fidelity technique would be an interesting addition to future work, though it is unclear how adulteration of the ambient gas composition (e.g., with tracer particles or smoke) will affect EAD device performance.

\section{Conclusion}
Understanding the influence of ground proximity effects on vehicle performance is critical for deploying them safely and efficiently in constrained environments. We show that there are significant effects to consider for small-scale electroaerodynamic propulsors depending on thruster geometry, multi-thruster arrangement, distance from the ground plane, and inclusion of exhaust flanges or fuselage strakes. These effects have the potential to either greatly improve efficiency in hover---for example, static thrust of a quadthruster was shown to increase by over 300$\%$ under certain conditions---or greatly reduce it---for example, static thrust was instead \textit{reduced} by 20$\%$ in only a slightly different configuration, at the same distance from the ground plane. Our results point towards fruitful areas of future investigation for development of a predictive model for these effects as they apply to EAD propulsion, and may open the door to power-autonomous ion-propelled vehicles designed to take advantage of proximity-mediated efficiency enhancement.



\section*{ACKNOWLEDGMENT}
The authors would like to thank the staff and sponsors of the University of Utah Undergraduate Research Opportunity Program (UROP) for supporting author G.N.. 


\bibliographystyle{IEEEtran}

\bibliography{prox}

\end{document}